\documentclass[11pt]{article}
\usepackage{amsmath, amssymb, amsfonts, amscd, xspace, pifont, natbib}
\usepackage{epsfig, amsfonts, verbatim, multirow}
\usepackage{url, algorithm, algorithmic, hyperref}
\usepackage{color}

\usepackage{graphicx}
\usepackage{caption}
\usepackage{subcaption}

\newcommand{\captionfonts}{\small}
\makeatletter  
\long\def\@makecaption#1#2{%
  \vskip\abovecaptionskip
  \sbox\@tempboxa{{\captionfonts #1: #2}}%
  \ifdim \wd\@tempboxa >\hsize
    {\captionfonts #1: #2\par}
  \else
    \hbox to\hsize{\hfil\box\@tempboxa\hfil}%
  \fi
  \vskip\belowcaptionskip}
\makeatother   

\newcommand{\V}[1]{\ensuremath{\boldsymbol{#1}}\xspace}

\newtheorem{theorem}{Theorem}

\begin{document}
\title{\Large \bf High-Dimensional Mixed Graphical Models}

\author{Jie Cheng$^{\dagger}$, Tianxi Li$^{\ddagger}$, Elizaveta Levina$^{\ddagger}$, Ji Zhu$^{\ddagger}$ \\ \normalsize $^\dagger$ Google, Inc.,   $^{\ddagger}$ 
Department of Statistics, University of Michigan \\}

\maketitle
\bigskip

\begin{abstract}

While graphical models for continuous data (Gaussian graphical models) and discrete data (Ising models) have been extensively studied, there is little work on graphical models for data sets with both continuous and discrete variables (mixed data), which are common in many scientific applications. We propose a novel graphical model for mixed data, which is simple enough to be suitable for high-dimensional data, yet flexible enough to represent all possible graph structures.   We develop a computationally efficient regression-based algorithm for fitting the model by focusing on the conditional log-likelihood of each variable given the rest.   The parameters have a natural group structure, and sparsity in the fitted graph is attained by incorporating a group lasso penalty, approximated by a weighted lasso penalty for computational efficiency. We demonstrate the effectiveness of our method through an extensive simulation study and apply it to a music annotation data set (CAL500), obtaining a sparse and interpretable graphical model relating the continuous features of the audio signal to binary variables such as genre, emotions, and usage associated with particular songs.   While we focus on binary discrete variables for the main presentation, we also show that the proposed methodology can be easily extended to  general discrete variables.

\end{abstract}
{\sc Key Words:} Conditional Gaussian density, Graphical model, Group lasso, Mixed variables, Music annotation.
	

\begin{center}
\section{Introduction} \label{sec:intro}
\end{center}
\setcounter{section}{1}
\setcounter{subsection}{0}

Graphical models have proven to be a useful tool in representing the conditional dependency structure of  multivariate distributions.   The undirected graphical model in particular, sometimes also referred to as the Markov network, has drawn a notable amount of attention over the past decade.  In an undirected graphical model, nodes in the graph represent the variables, while an edge between a pair of variables indicates that they are dependent conditional on all other variables.   The vast majority of the graphical models literature has been focusing on either the multivariate Gaussian model (see \citet{meinshausen06, Yuan07, spice, banerjee06, Rocha08, Ravikumar08, Lam07,  Peng09, yuan10, Cai2011, fht08}), or the Ising model for binary and discrete data (see \citet{Hofling09, Ravikumar10}).  The properties of these models are by now well understood and studied both in the classical and the high-dimensional settings. Both these models can only deal with variables of one kind -- either all continuous variables in Gaussian models or all binary variables in the Ising model (extensions of the Ising model to general discrete data, while possible in principle, are rarely used in practice).   In many applications, however, data sources are complex and varied, and frequently result in mixed types of data, with both continuous and discrete variables present in the same dataset.  In this paper, we will focus on graphical models for this type of  mixed data (mixed graphical models).   

Sparse estimation of Gaussian graphical models using regularized maximum likelihood methods using $\ell_1$ penalty on the precision matrix has received a lot of attention in recent years. \citet{fht08} developed an efficient algorithm known as the graphical lasso, with excellent theoretical properties and fast implementations available, but its reliance on the assumption of normality can be restrictive in many real applications.  \citet{Liu2009} relaxed this assumption to a Gaussian copula model they called nonparanormal. The authors assume that there exist differentiable, monotone transformations $f = (f_1, f_2, \ldots, f_p)$ such that, $f(X) = (f_1(X_1), f_2(X_2), \ldots, f_p(X_p))$ is Gaussian with mean $\mu^f$ and precision matrix $\Omega^f$. Then $X_i$ and $X_j$ are conditionally independent given the rest if and only if $\Omega^f_{ij} = 0$. The proposed algorithm estimates the marginal transformation functions non-parametrically and applies graphical lasso algorithm on the transformed data to estimate the underlying graphical structure.  \citet{Liu2012} and \citet{Xue2012} independently exploited the connection of non-parametric rank based correlation estimators such as Spearman's rho \citep{Spearman1904} and Kendall's tau \citep{Kendall1938} with the nonparanormal covariance matrix to directly estimate $\Omega^f$ avoiding the plug-in estimator involving $\hat{f_j}$. \citet{Liu2012} established that their proposed estimator 
achieves optimal rates of convergence for graph recovery and parameter estimation, while \citet{Xue2012} investigated the theoretical properties of rank-correlation based algorithms including graphical lasso, Dantzig selector \citep{Candes2007} and  CLIME \citep{Cai2011}. This line of work has clearly extended the scope of graphical models to general continuous data but they are not suitable for binary variables and therefore cannot handle mixed type of variables, which is the main setting of this paper. 

For binary data, most of the literature has focused on the Ising model \citep{Ising1925}, originally proposed in statistical physics. For high-dimensional data, the Ising model becomes computationally challenging due to the intractability of the log partition function. \citet{Ravikumar10} used an $\ell_1$-penalized pseudo-likelihood method to estimate the edge set, in the spirit of the neighborhood selection method proposed by \citet{meinshausen06} for continuous data. More recently, \citet{Xue2012b} proposed using a SCAD penalty \citep{Fan2001} to estimate a sparse graphical model;  they developed scalable and efficient algorithms for optimizing the non-concave problem and proved theoretical performance guarantees superior to concave penalties such as the lasso. 

For mixed data containing both continuous and discrete variables, the conditional Gaussian distribution  \citep{Lauritzen89, Lauritzen96}  has become the foundation of most developments on this topic. In the original paper, \citet{Lauritzen89} defined a general form of the conditional Gaussian density and characterized the connection between the model parameters and the conditional associations among the variables. The model is fitted via the maximum likelihood approach. The number of parameters in this model, however, grows exponentially with the number of variables, which renders it unsuitable for high-dimensional problems arising in many modern  applications. \citet{Edwards90} generalized the conditional Gaussian distribution model to the hierarchical interaction model which can account for all possible hierarchical/nested interactions between the discrete and continuous variables, and proposed methods of maximum likelihood estimation under these models using marginal mean and covariance calculations. They also showed some connection of hierarchical interaction models with MANOVA type of models.  Much more recently, \citet{Lee12} and \citet{Fellinghauer11} have studied the mixed graphical model (simultaneously and independently of the present paper), under a setting that could be viewed as a simplified special case of our proposal. \citet{Edwards10} also proposed an extended algorithm based on the Chow-Liu algorithm \citep{ChowLiu68}   for the multivariate discrete case to fit high-dimensional mixed graphical models.  A more detailed discussion of these papers is postponed to Section \ref{sec:summary}.  


In this paper, we propose a simplified version of the conditional Gaussian distribution which reduces the number of parameters significantly yet maintains flexibility. To fit the model in a high-dimensional setting, we impose a sparsity assumption on the underlying graph  and develop a node-based regression approach with the group lasso penalty \citep{Yuan06}, since edges in the mixed graphical model are associated with groups of parameters.   The group lasso penalty in itself is not computationally efficient due to the overlaps between groups, and we develop a much faster weighted $\ell_1$ approximation to the group penalty which is of independent interest. The simulation results show promising model selection performance in terms of estimating the true graph structure under high-dimensional settings.

We start with a brief introduction to conditional Gaussian distribution and its Markov properties following \citet{Lauritzen96}.

\textbf{Conditional Gaussian (CG) density}: let $X = (Z, Y)$ be a mixed random vector, where  $Z = (Z_j)_{j \in \Delta}$ is a $q$-dimensional discrete sub-vector, $Y = (Y_\gamma)_{\gamma \in \Gamma}$ is a $p$-dimensional continuous sub-vector, and $\Delta$ and $\Gamma$ are index sets for $Z$ and $Y$, respectively.
The conditional Gaussian density $f(x)$ is defined as 
\begin{equation}\label{CGdensity}
f(x) = f(z, y) = \exp\left(g_z + h_z^Ty - \frac{1}{2}y^TK_zy\right),
\end{equation}
where $\{(g_z, h_z, K_z), g_z \in \mathbb{R}, h_z \in \mathbb{R}^p, K_z \in \mathbb{R}^+_{p\times p}\}$ 
are the canonical parameters of the distribution. The following equations connect the canonical parameters in \eqref{CGdensity} to the moments of $Y$ and $Z$:   
\begin{align} \label{moments}
 &P_z = P(Z = z) = (2\pi)^{p/2}(\det(K_z))^{-1/2}\exp\left(g_z + h_z^TK_z^{-1}h_z/2\right), \notag\\ 
&\xi_z = \mathbb{E}(Y \lvert Z = z) = K_z^{-1}h_z,  \notag\\
&\Sigma_z = \mathrm{Var}(Y\lvert Z = z) = K_z^{-1} . 
\end{align}
Also, $\mathcal{L}(Y\lvert Z=z) = \mathcal{N}(\xi_z, \Sigma_z)$, so conditional on $Z = z$, each $Y$ is normally distributed with the mean and variance determined by $z$.  
The next theorem relates the graphical Markov property of the model to its canonical parameters and serves as the backbone of the subsequent analysis.\newline

\begin{theorem} \label{thm:CGMarkov} \citep{Lauritzen89}
Represent the canonical parameters from \eqref{CGdensity} by the following expansions, 
\begin{equation} \label{expansion}\displaystyle
g_z = \sum_{d:d\subseteq \Delta}\lambda_d(z), \quad h_z = \sum_{d:d\subseteq \Delta}\eta_d(z), \quad K_z = \sum_{d:d\subseteq \Delta}\Phi_d(z) \ , 
\end{equation}
where functions indexed by the index set $d$ only depend on $z$ through $z_d$.
 Then a CG distribution is Markovian with respect to a graph $\mathcal{G}$ if and only if the density has an expansion that satisfies 
\begin{eqnarray*}
\lambda_d(z) &\equiv& 0 \qquad \mbox{unless $d$ is complete in $\mathcal{G}$}, \\
\eta_d^{\gamma}(z) &\equiv& 0 \qquad \mbox{unless $d\cup\{\gamma\}$ is complete in $\mathcal{G}$},\\
\Phi_d^{\gamma\mu}(z) &\equiv& 0 \qquad \mbox{unless $d\cup\{\gamma, \mu\}$ is complete in $\mathcal{G}$}.
\end{eqnarray*}
where $\eta_d^{\gamma}(z)$ is the $\gamma$-th element of $\eta_d(z)$, $\Phi_d^{\gamma\mu}(z)$ is the $\gamma\mu$-th element of $\Phi_d(z)$, and a subgraph is called complete if it is fully connected. 
\end{theorem}

The rest of the paper is organized as follows.  Section \ref{sec:model} introduces the simplified mixed graphical model which has just enough parameters to cover all possible graph structures and proposes an efficient estimation algorithm for the model. Section \ref{sec:simulation} uses several sets of simulation studies to evaluate the model selection performance and compare to some alternative methods for graph estimation.  In Section \ref{sec:realdata}, the proposed model is applied to a music annotation data set \textit{CAL500} with binary labels and continuous audio features.  In Section \ref{sec:extension}, we describe the generalization of the model from binary to  discrete variables.  Finally, we conclude in Section \ref{sec:summary} with a discussion.

\begin{center}
\section{Methodology} \label{sec:model}

\end{center}
\setcounter{section}{2}
\setcounter{subsection}{0}
We propose a simplified but flexible version of the conditional Gaussian model for mixed data.  The model fitting  is based on maximizing the conditional log-likelihood of each variable given the rest, for computational tractability. This leads to penalized regression problems with overlapping groups of  parameters. The natural solution to the problem is to fit separate regressions with an overlapping group lasso penalty.    This is computationally quite expensive, so we approximate the overlapping group lasso penalty by an appropriately weighted  $\ell_1$ penalty.

\subsection{The simplified mixed graphical model}\label{secsec:simplemodel}
Without loss of generality, we partition the random vector X = $(Z_1, Z_2, \ldots, Z_q, Y_1, Y_2, \ldots, Y_p)$ into the binary part with $\Delta = \{1,2, \cdots, q\}$ and the continuous part with $\Gamma = \{1,2, \cdots, p\}$.  We propose to consider the conditional Gaussian distribution with the density function
\begin{align}
\log  & f(z, y) = \sum_{d:d\subseteq\Delta, \lvert d\rvert \leq 2} \lambda_d(z) + \sum_{d:d\subseteq\Delta, \lvert d\rvert \leq 1} \eta_d(z)^Ty - \frac{1}{2}\sum_{d:d\subseteq\Delta, \lvert d\rvert \leq 1} y^T\Phi_d(z)y  \nonumber \\ 
&= \big(\lambda_0 + \sum_j \lambda_j z_j + \sum_{j > k} \lambda_{jk}z_jz_k\big) + 
y^T\big(\eta_0 + \sum_j \eta_j z_j \big) 
- \frac{1}{2}y^T\big(\Phi_0 + \sum_{j=1}^q\Phi_j z_j \big)y \nonumber  \\ 
&= \left(\lambda_0 + \sum_j \lambda_j z_j + \sum_{j > k} \lambda_{jk}z_jz_k\right) + \sum_{\gamma = 1}^p \left(\eta_0^{\gamma} + \sum_j \eta_j^{\gamma}z_j \right)y_{\gamma}  \nonumber \\ 
& \ \ \ \  \ - \frac{1}{2}\sum_{\gamma, \mu = 1}^p\left(\Phi_0^{\gamma\mu} + \sum_{j=1}^q\Phi_j^{\gamma\mu}z_j \right)y_{\gamma}y_{\mu}  \ , 
\label{model}
\end{align}
where $\{\textrm{diag}(\Phi_j)\}_{j=1}^q = \{\Phi_j^{\gamma\gamma}; j = 1, \ldots, q , \  \gamma = 1, \ldots, p\}$ are all 0 
and $\lambda_0$ is the normalizing constant, 
\begin{equation*}\displaystyle
\lambda_0^{-1} = (2\pi)^{\frac{p}{2}}\sum_{z\in\{0, 1\}^q}\det(K_z)^{\frac{1}{2}} \exp\left( \sum_j\lambda_jz_j + \sum_{j > k} \lambda_{jk}z_jz_k + \frac{ h_z^TK_z^{-1}h_z}{2}\right).
\end{equation*}
Note that the density is explicitly defined via the expanded terms in \eqref{expansion} and the canonical parameters $(g_z, h_z, K_z)$ can be obtained immediately by summing up the corresponding terms. This model simplifies the full conditional Gaussian distribution \eqref{CGdensity} in two ways:  first, it omits all interaction terms between the binary variables of order higher than two, and second, it models the conditional covariance matrix and the canonical mean vector of the Gaussian variables as a linear function of the binary variables instead of allowing arbitrary dependence on the  binary variables.   These simplifications  reduce the total number of parameters from $\mathcal{O}(p^22^{(p+q)})$ in the full model to $\mathcal{O}\left(\max(q^2, p^2q)\right)$. 
This reduction is necessary especially in the high-dimensional setting, where there are limited number of samples and even if the true model involves higher order interactions, it may not be possible to estimate them well due to the bias-variance trade-off.
On the other hand, this model is the simplest CG density, among those allowing for varying conditional covariance $\textrm{Var}(Y\vert Z)$ , that can represent all possible graph structures, since it includes interactions between all the continuous and discrete variables and thus allows for a fully connected graph, an empty graph, and everything in between.  The fact that it allows both the conditional mean and the conditional covariance of $Y$ given $Z$ to depend on $Z$ adds flexibility.

\subsection{Parameter Estimation}
Given sample data $\{(\V{z}_i, \V{y}_i)\}_{i=1}^n$, directly maximizing the log-likelihood $\sum_{i=1}^n \log f(\V z_i, \V y_i)$ is impractical due to the normalizing constant $\lambda_0$. The conditional likelihood of one variable given the rest, however, is of much simpler form and easy to maximize.   Hence, we focus on the conditional log-likelihood of each variable and fit separate regressions to estimate the parameters, much in the spirit of the neighborhood selection approach proposed by \citet{meinshausen06} for the Gaussian graphical model and by \citet{Ravikumar10} for the Ising model.    To describe the conditional distributions, let $Z_{-j} = (Z_1, \ldots,Z_{j-1}, Z_{j+1}, \ldots, Z_q )$ and  $Y_{-\gamma} = (Y_1, \ldots,Y_{\gamma-1}, Y_{\gamma+1}, \ldots, Y_p )$. Then the conditional distribution of $Z_j$ given $(Z_{-j}, Y)$ is described by 
\begin{equation} \label{logistic_regression} 
\displaystyle\log\frac{P(Z_j = 1 \lvert Z_{-j}, Y)}{P(Z_j = 0 \lvert Z_{-j}, Y)} = \lambda_j + \sum_{k\neq j}\lambda_{jk}Z_k + \sum_{\gamma = 1}^p \eta_j^{\gamma}Y_{\gamma} - \frac{1}{2}\sum_{\gamma, \mu = 1}^p \Phi_j^{\gamma\mu} Y_{\gamma}Y_{\mu} \ .
\end{equation}
Since the conditional log-odds in \eqref{logistic_regression} is linear in parameters, maximizing this conditional log-likelihood can be done via fitting a logistic regression with $(Z_{-j}, Y, Y^2)$ as predictors and $Z_j$ as response. 

For the continuous variables, the conditional distribution of $Y_\gamma$ given $(Y_{-\gamma}, Z)$ is given by 
\begin{equation*}
\displaystyle Y_{\gamma} =\frac{1}{K_z^{\gamma\gamma}} \left( \eta_0^{\gamma} + \displaystyle\sum_j \eta_j^{\gamma}Z_j  - \sum_{\mu \neq \gamma}\left(\Phi_0^{\gamma\mu} + \sum_j\Phi_j^{\gamma\mu}Z_j \right)Y_{\mu} \right)
+ e_{\gamma},
\end{equation*}
 where $e_{\gamma} \sim \mathcal{N}\left(0, (K_z^{\gamma \gamma})^{-1}\right)$. With $\textrm{diag}(\Phi_j) =0 $ as defined by \eqref{model}, we have $K_z^{\gamma \gamma} = \Phi_0^{\gamma\gamma}$, i.e., the conditional variance of $Y_{\gamma}$ does not depend on $Z$.  Rewrite
\begin{equation} \label{linear_regression}\displaystyle
Y_{\gamma} = \tilde\eta_0^{\gamma} + \sum_j \tilde\eta_j^{\gamma}Z_j - \sum_{\mu \neq \gamma} \left(\tilde\Phi_0^{\gamma\mu} + \sum_j\tilde\Phi_j^{\gamma\mu}Z_j \right)Y_{\mu} + e_{\gamma} \ , 
\end{equation}
where the redefined parameters with ``tilde'' are proportional to the original ones up to the same constant for each regression. Again, the conditional mean of $Y_\gamma$ is linear in parameters, which can be estimated via ordinary linear regression with predictors $(Y_{-\gamma}, Z, Y_{-\gamma}Z)$ and response $Y_\gamma$. 

\subsection{Regularization}
Based on Theorem \ref{thm:CGMarkov},  the following equivalences hold:
\begin{eqnarray}\label{eq:equivalence}
Z_j \perp Z_k \  \vert \ \ X \backslash \{Z_j, Z_k\} &\Longleftrightarrow & \lambda_{jk} = 0,  \nonumber \\
Z_j \perp Y_{\gamma} \ \vert \ \ X \backslash\{Z_j, Y_\gamma\} &\Longleftrightarrow& \V \theta_{j\gamma} = \left(\eta_j^{\gamma}, \{\Phi_j^{\gamma\mu}: \mu \in \Gamma\backslash\{\gamma\}\} \right) = 0,  \nonumber \\ \label{edge_parameter}
Y_{\gamma} \perp Y_{\mu} \ \vert \ \ X\backslash\{Y_\gamma, Y_\mu\} &\Longleftrightarrow& \V \theta_{\gamma\mu} = \left(\Phi_0^{\gamma\mu}, \{\Phi_j^{\gamma\mu}: j \in \Delta\}\right) =  0.
\end{eqnarray}
This means that each edge between pairs of $(Z_j, Y_\gamma)$ and $(Y_\gamma, Y_\mu)$ depends on a parameter vector, denoted by $\V{\theta}_{j\gamma}$ and $\V \theta_{\gamma\mu}$, respectively.  
To encourage sparsity of the edge set under high-dimensional settings, we add the $\ell_1\backslash\ell_2$ penalty, proposed by \citet{Yuan06} for group lasso, to the loss function in each regression.  The groups are pre-determined by parameter vectors corresponding to each edge. Denoting the loss function for the logistic regression of $Z_j$ by $\ell_j$ and the linear regression for $Y_\gamma$ by $\ell_\gamma$, we have 
\begin{eqnarray*}
\ell_j &=& - \frac{1}{n} \displaystyle \sum_{i=1}^n \log (P(z_{ij} \ | \ (\V z_{i,(-j)}, \V y_i)), \\
\ell_\gamma &=& \frac{1}{n}\displaystyle \sum_{i=1}^n( y_{i\gamma} - ( \tilde\eta_0^{\gamma} + \sum_{j=1}^q \tilde\eta_j^{\gamma}z_{ij} - \sum_{\mu \neq \gamma} (\tilde\Phi_0^{\gamma\mu} + \sum_{j=1}^q\tilde\Phi_j^{\gamma\mu}z_{ij} )y_{i\mu}))^2.
\end{eqnarray*}
We estimate the parameters by optimizing the following criteria separately, for $j = 1,\ldots, q$ and $\gamma = 1, \ldots, p$ 
\begin{eqnarray} \label{logistic_regression_l1l2} \displaystyle
\textrm{\textit{Logistic regression}: }  \min \ \ell_j &+& \rho \left( \kappa \sum_{k\neq j}\vert\lambda_{jk}\vert + \sum_{\gamma=1}^p \|\V \theta_{j\gamma}\|_2 \right),  \\ \label{linear_regression_l1l2}
\textrm{\textit{Linear regression}: } \min \ \ell_\gamma &+& \rho \left( \sum_{\mu \neq \gamma}\|\tilde{\V \theta}_{\gamma\mu}\|_2 +\sum_{j=1}^q \|\tilde{\V\theta}_{j\gamma}\|_2 \right) \ , 
\end{eqnarray}
where $\rho$ and $\kappa$ are tuning parameters.  Using two tuning parameters, $\rho$ and $\kappa$, allows us to penalize individual parameters and groups of parameters differently, essentially allowing the edges between binary variables to be penalized differently from other edges.  While in principle both parameters can be tuned, in simulations we got good and very stable results over the range $0.1 \le \kappa \le 0.5$, and thus in simulations we set $\kappa=0.1$.  
Note that we use the same tuning parameter $\rho$ for both linear and logistic regressions.  One reason to use a single tuning parameter $\rho$ is to simplify the treatment of overlapping groups of parameters from different regressions (see more on this below).  We did conduct simulation experiments using two different tuning parameters,  $\rho_1$ for linear regression and $\rho_2$ for logistic regression, and the results (not shown in this paper) are similar to simply using one tuning parameter $\rho$ for both linear and logistic regressions.  In principle, one could also tune each regression separately, but the computational cost is prohibitive; further the estimation variance can be extraordinarily high when using a large number of tuning parameters, see for example \citet{meinshausen06} and \citet{Ravikumar10} for similar neighborhood selection settings where a single tuning parameter was used for different regressions.
Finally, note that in linear regression, the parameters in \eqref{linear_regression} denoted with ``tilde'' are proportional to the original parameters.   The original parameters can be recovered by multiplying the estimates by $(\hat{K}_z^{\gamma\gamma})^{-1}$, which can be estimated from the mean squared error of the linear regression. 

Although the optimization problems \eqref{logistic_regression_l1l2} and \eqref{linear_regression_l1l2} appear to be group lasso regressions, they cannot be solved by regular group lasso algorithms, because the groups of parameters involved in each regression overlap.   Specifically, in logistic regression, parameter $\Phi_{j}^{\gamma\mu}$ is part of both  $\V \theta_{j\gamma}$ and $\V \theta_{j\mu}$ and affects both the edges $(Z_j, Y_\gamma)$ and $(Z_j, Y_\mu)$; thus $\V \theta_{j\gamma}$ has one parameter overlapping with each of the other $\V \theta_{j\mu}$'s.  Similarly, in linear regression, $\Phi_j^{\gamma\mu}$ is part of both $\V \theta_{j\gamma}$ and $\V \theta_{\gamma\mu}$, and affects both the edges $(Z_j, Y_\gamma)$ and $(Y_\gamma, Y_\mu)$. This overlapping pattern creates additional difficulties in using the group penalty to perform edge selection. The overlapping group lasso problem was theoretically investigated by \citet{jenatton2011structured} (see also \citet{jacob2009group}) but has received limited attention from a computational point of view. \citet{Yuan11} recently proposed an algorithm for solving the overlapping group lasso problem but the speed is still an issue that limits its capability of handling high-dimensional data. Therefore we took the approach of finding a surrogate for the overlapping group lasso penalty to make the problem both efficient to solve and suitable for high-dimensional settings without losing much accuracy. 

\begin{figure}[h!]
\begin{center}
\includegraphics[height = 4in, width = 5in]{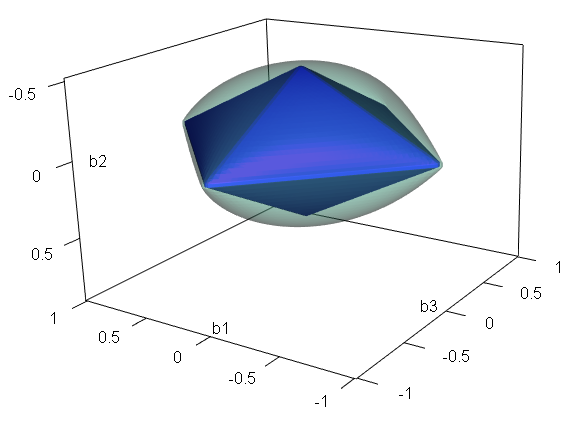}
\vspace{-0.3in}
\caption{\textit{Green (outside)}: $\{\V b: \sqrt{b_1^2 + b_2^2} + \sqrt{b_3^2 + b_2^2} = 1\}$; \textit{Blue (inside)}:  $\{\V b: |b_1| + |b_3| + 2|b_2| = 1\}$}
\label{fig:contour}
\end{center}
\end{figure}

Instead of the overlapping group lasso penalty, we propose to use its upper bound as a surrogate, which is essentially a weighted $\ell_1$ penalty. The upper bound results from the fact that for any vector $\V b$, $\|\V b\|_2 \leq \|\V b\|_1$. Take the logistic regression \eqref{logistic_regression_l1l2} for example, $ \sum_{k\neq j}\vert\lambda_{jk}\vert + \sum_{\gamma=1}^p \|\V \theta_{j\gamma}\|_2 \leq \sum_{k\neq j}\vert\lambda_{jk}\vert + \sum_{\gamma=1}^p\vert \eta_j^\gamma\vert + 2\sum_{\gamma < \mu}\vert\Phi_j^{\gamma\mu}\vert $. The surrogate on the right penalizes the overlapped parameters twice as much as the other parameters, which makes intuitive sense since incorrectly identifying the overlapped parameters as non-zero will result in two wrong edges, while the incorrect unique parameters for  each group will only cause one wrong edge.

To illustrate the upper bound geometrically, we show a toy example. Suppose the parameter vector is $\V b = (b_1, b_2, b_3)$, and two groups are $\mathcal{G}_1 = (b_1, b_2)$ and $\mathcal{G}_2 = (b_2, b_3)$. The optimization problem for the overlapping group lasso penalty and its $\ell_1$ surrogate boils down to optimizing the same loss function over different feasible regions (for the same tuning parameter). Figure \ref{fig:contour} compares the feasible regions $\mathcal{R}_1 = \{\V b: \sqrt{b_1^2 + b_2^2} + \sqrt{b_3^2 + b_2^2} \leq 1\}$ and $\mathcal{R}_2 = \{\V b: |b_1| + |b_3| + 2|b_2| \leq 1\}$. Since both the logistic loss and the least squares loss are smooth convex functions, their optima are likely to occur at singular points of the feasible region. Note that $\mathcal{R}_2$ is not only a subset of $\mathcal{R}_1$ but it contains all four singular points of $\mathcal{R}_1$: $(\pm 1, 0, 0), (0,0,\pm 1)$. Thus for this example, it is guaranteed that all the optimal points of singular points on $\mathcal R_1$ for the overlapping group lasso penalty are also optimal points for its surrogate. 
The effectiveness of this approximation will be further demonstrated by a simulation study in Section~\ref{sec:simulation}.

With the penalty being replaced by the weighted $\ell_1$ surrogate, we solve the following regression problems separately as an approximation to the original problems $\eqref{logistic_regression_l1l2}$ and \eqref{linear_regression_l1l2} to obtain the parameter estimates. 

\noindent \textit{Logistic regression with $\ell_1$ penalty}: for $j = 1, \ldots , q$ 
\begin{equation} \label{logistic_regression_l1} \displaystyle
\min \ \ell_j + \rho \left( \kappa \sum_{k\neq j}\vert\lambda_{jk}\vert +  \sum_{\gamma=1}^p\vert \eta_j^\gamma\vert + 2\sum_{\gamma < \mu}\vert\Phi_j^{\gamma\mu}\vert \right).  
\end{equation}
\textit{Linear regression with $\ell_1$ penalty}: for $\gamma = 1, \ldots, p$
\begin{equation}\label{linear_regression_l1} \displaystyle
\min \ \ell_\gamma +\rho \left( \sum_{j=1}^q\vert \tilde{\eta}_j^\gamma\vert + \sum_{\mu \neq \gamma} \vert \tilde{\Phi}_0^{\gamma\mu} \vert + 2\sum_{j=1}^q\sum_{\mu \neq \gamma}\vert \tilde{\Phi}_{j}^{\gamma\mu}\vert\right).
\end{equation}
Since we are estimating  parameters in separate regressions, all parameters determining edges will be estimated at least twice.   This situation is common in all neighborhood selection approaches based on separate regressions, and is usually solved by taking either the largest or the smallest (in absolute value) of the estimates.   Here we chose taking the maximum of absolute values as the final estimate, based on simulations studies (not shown) which resulted in the maximum giving better model selection results than the minimum.  To fit both types of regressions with a weighted $\ell_1$ penalty, we used the matlab package \textit{glmnet} of \citet{Friedman&Hastie&Tibshirani08_glmnet}.  
%
%

\begin{center}
\section{Numerical performance evaluation}\label{sec:simulation}
\end{center}
\setcounter{section}{3}
\setcounter{subsection}{0}
In this section, we first demonstrate the effectiveness of the weighted lasso approximation to the overlapping group lasso.  Then we show simulation results regarding model selection performance under different settings and comparison with other graph selection methods \citep{Lee12,Fellinghauer11}. The results for graph selection are summarized in ROC curves, where we plot the true positive rate (TPR) against the false positive rate (FPR), for both parameters and edges across a fine grid of tuning parameters.  All ROC curves are obtained based on 100 replications with local smoothing.  Let $\V\theta$ and $\hat{\V\theta}$ denote the true parameter vector and the fitted parameter vector respectively (without the intercept terms in the regressions).  The parameter based quantities of interest are defined as
\begin{eqnarray*}
\textrm{TP} &=& \# \{j:\hat{\theta}_j \neq 0 \ \textrm{and} \ \theta_j \neq 0\},\ \ \   \textrm{FP} = \# \{j:\hat{\theta}_j \neq 0 \ \textrm{and} \ \theta_j = 0\},\\
\textrm{TPR} &=& \frac{\textrm{TP}}{\# \{j: \theta_j \neq 0\}}, \ \ \ \ \ \ \ \ \ \ \ \ \ \ \ \textrm{FPR} = \frac{\textrm{FP}}{\#\{j: \theta_j = 0\}} . 
\end{eqnarray*}
The quantities based on the true edge set $\V E$ and the estimated edge set $\hat{\V E}$ can be defined in a similar fashion. \newline

\subsection{Weighted lasso approximation to the overlapping group lasso}\label{secsec:approximation}
 
We first briefly discuss the weighted lasso approximation to the overlapping group lasso penalty in the setting of linear regression, since this approximation itself is not limited to graphical models.   One of the most frequent uses of group lasso in regression is to encourage a hierarchical variable selection path \cite{zhao2009composite}, to ensure main effects are selected before the corresponding interactions are included.   \citet{zhao2009composite} designed an example to investigate this property, and here we use the same setting to compare the proposed weighted lasso approximation as well as the regular lasso to the overlapping group penalty.   Following \citet{zhao2009composite} exactly, we have variables $x_1, \dots, x_{10}$ and all of their pairwise products $x_ix_j, 1\le i < j \le 10$ in the model,  resulting in 55 variables. 
 The variables are generated from a standard normal distribution; the coefficients of the first four variables are $7, 2, 1, 1$, and all main effects and interaction effects involving the other six are zero.   The response follows the model $Y = X\beta + \epsilon$, where $\epsilon \sim N(0, 3.7 I)$, and the sample size $n=121$.    We consider three settings of \citet{zhao2009composite} for the interaction effects between $x_1, \cdots, x_4$, shown in Table~\ref{tab:ZhaoConfig}, corresponding to weak, moderate, and strong interaction effects;  we omit their two other cases to save space (the results are similar).  The groups we use include $\{x_i, x_ix_j, j\ne i\}, i = 1, \cdots, 10$ and singleton groups with only one interaction term $\{x_ix_j\}, i\ne j$. So there are 10 groups of size 10 and 45 groups of size 1.

\begin{table}
\centering
\begin{tabular}{l|rrrrrr}
  \hline
 & $x_1x_2$ & $x_1x_3$ & $x_1x_4$ & $x_2x_3$ & $x_2x_4$ & $x_3x_4$ \\ 
  \hline
Weak & 1 & 0 & 0 & 0.5 & 0.4 & 0.1 \\ 
Moderate & 5& 0 & 0 & 4 & 2 & 0\\ 
Strong & 7 & 7 & 7 & 2 & 2 &1 \\ 
   \hline
\end{tabular}
\caption{Three configurations of interaction effects (weak, moderate and strong) from  \citet{zhao2009composite}.}
\label{tab:ZhaoConfig}
\end{table}

\begin{figure}
  \centering
  \begin{subfigure}{\linewidth}
    \centering
    \includegraphics[width=1\linewidth]{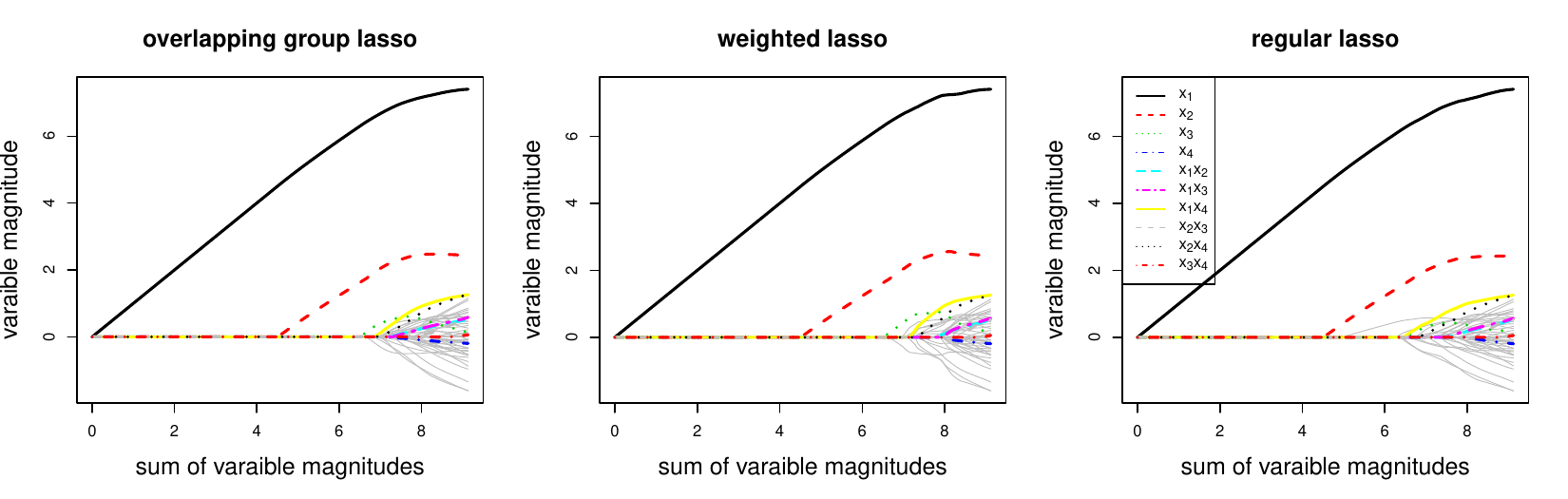}
    \caption{Weak interactions. }
  \end{subfigure}
  \begin{subfigure}{\linewidth}
    \centering
    \includegraphics[width=1\linewidth]{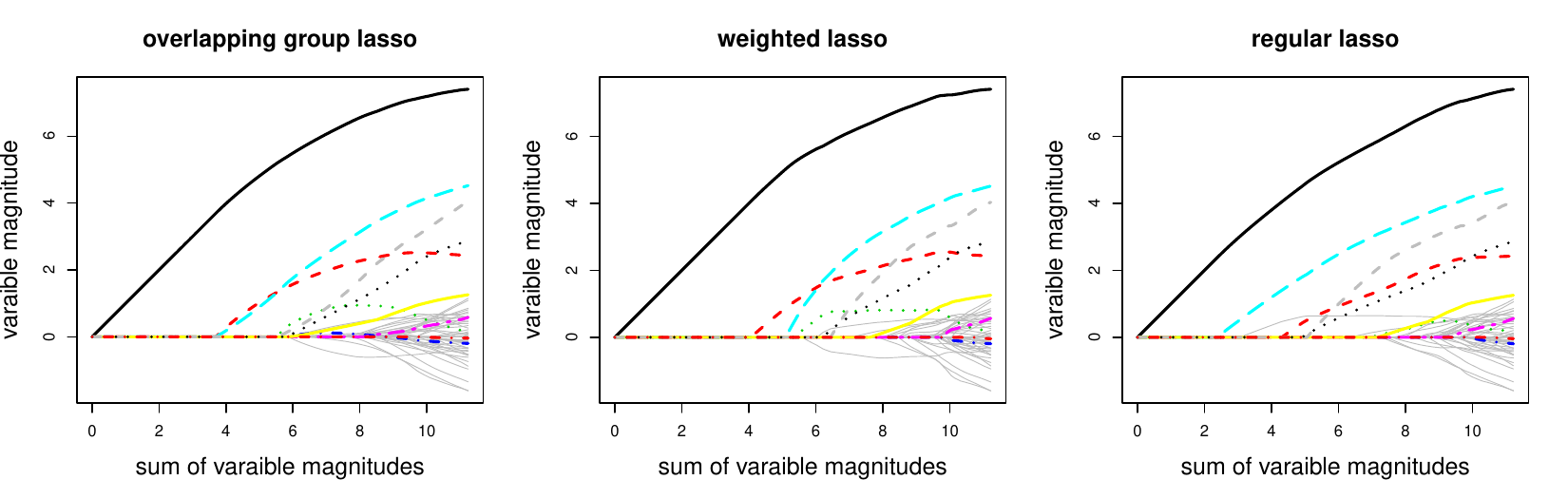}
    \caption{Moderate interactions.}
  \end{subfigure}  
 \begin{subfigure}{\linewidth}
    \centering
    \includegraphics[width=1\linewidth]{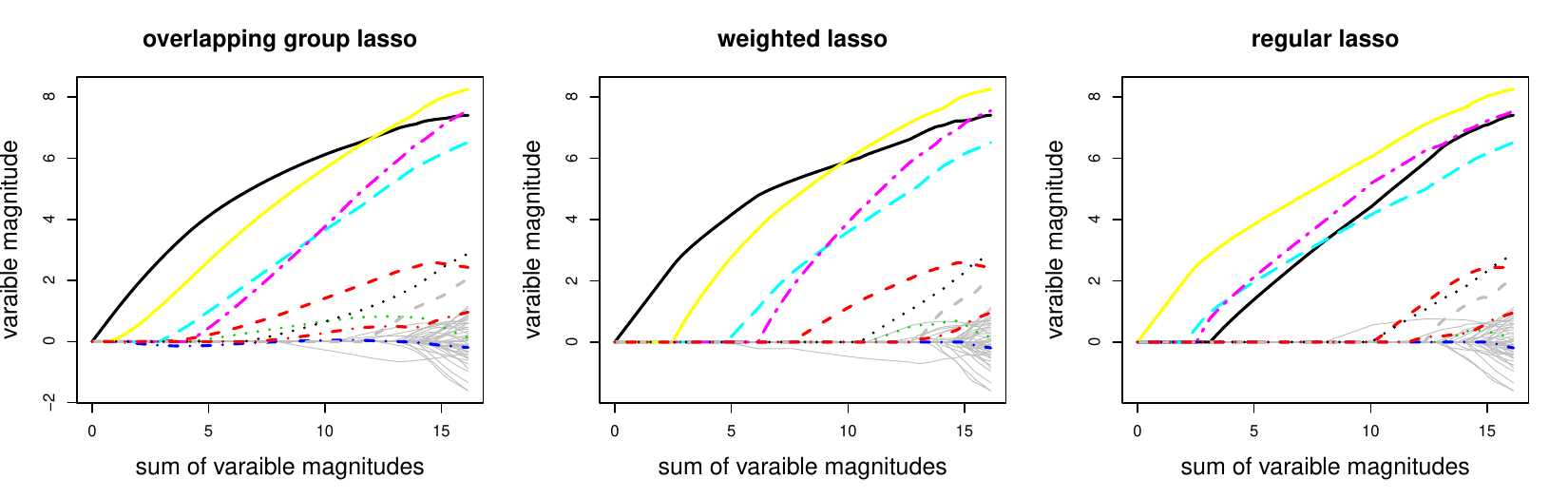}
    \caption{Strong interactions.}
  \end{subfigure}  
  \caption{ \label{fig:ZhaoVar}Variable selection paths of overlapping group lasso, weighted lasso and regular lasso. The x-axis is $\sum_i|{\beta}_i|$ and the y-axis is the value of the coefficients.}
\end{figure}

Figure~\ref{fig:ZhaoVar} shows the variable selection path for overlapping group lasso, weighted lasso and regular lasso for the three settings.  We focus on the four active variables and their  pairwise interactions, shown in bold colored curves in Figure~\ref{fig:ZhaoVar}; the remaining 45 coefficient paths are shown in thin gray curves.  For weak interactions, the three methods give nearly identical results. This is expected since the interaction effects are too weak to make any difference and without the interaction effects, the three penalties are the same. As the interaction effects become stronger, the difference becomes clear.    Note that the nature of the overlapping group lasso may let a subset of variables enter the model simultaneously, but this is typically not true for weighted lasso.   Thus it is reasonable to treat the lasso approximation as correct as long as all the variables in such a subset enter the model before any others.  For moderate effects, the weighted lasso gives the same variable selection as the overlapping group lasso as well.  For strong effects, weighted lasso makes mistakes on the two weakest main effects,  $x_3$  and $x_4$ (the green and blue dotted curves) and $x_4$ (the blue dotted curve), which remain close to zero along the entire group lasso path.  The regular lasso differs from the overlapping group lasso on $x_3$ for moderate interactions, and makes even more mistakes for strong interactions, missing even the strongest effect $x_1$ (black curve).

Overall, in this example the weighted lasso approximates the overlapping group lasso well, and much better than regular lasso, except when the interaction effects are weak.   In general settings, the quality of the weighted lasso approximation to the overlapping group lasso can depend on many factors, such as the pattern and degree of overlap between groups, the signal to noise ratio, the degree of correlation between predictors, etc.    Further investigation of this topic in the context of regression is beyond the scope of this paper and is left for future work.

\subsection{Parameter estimation and edge identification}
\label{secsec:performance}
%

To start, we investigate the impact of heterogeneity of node degrees (i.e., the number of edges connected to a node) of the underlying graph on the performance of the proposed method, as it is generally a challenge for graphical models.  We set the first $q = 10$ variables to be binary and the remaining $p = 90$ variables to be continuous, with the sample size $n = 100$. We first vary the maximum node degree by setting it to be 2, 6, and 10 in the graph while maintaining the total number of edges fixed at 80.    The smaller the maximum node degree is, with fixed total number of edges, the more homogeneous the degree distribution.

We generate the graph using the Erdos-Renyi model, and simply use rejection sampling to enforce the constraints, i.e., we keep generating the graph until the maximum node degree meets the requirement.  The edges of the graph are of three types: edges connecting binary variables (ZZ), edges connecting continuous variables (YY), and edges connecting binary and continuous variables (ZY).    In order to be able to compute the true positive rates for each category, we further require the graph have at least one edge in each category. 

Once the graph is fixed, we set all parameters corresponding to absent edges to 0.  
For the non-zero parameters, we set $\{\lambda_j, \lambda_{jk}, \eta_j\}$ to be positive or negative with equal probability and the absolute value of each non-zero $\eta_j$ is drawn from the uniform distribution on the interval $(0.9a, 1.1a)$ and each non-zero $\lambda_j$ or $\lambda_{jk}$ is from $(0.9c,1.1c)$.   For $\{\Phi_0, \Phi_j\}$, we set the off-diagonal elements to be positive or negative with equal probability, and the absolute value of each non-zero parameter is also drawn from a uniform distribution, on the interval $(0.9b, 1.1b)$.   The parameters $a$, $b$ and $c$ control the overall magnitude of the non-zero parameters and therefore the effective signal-to-noise ratio;  we set $a=c=1$ and $b=2$.    For the purpose of investigating the effect of the maximum node degree, varying the values of $a,b$ and $c$ does not result in a qualitative difference in the results.   The diagonal elements of $\Phi_0$ are chosen so that $\Phi_0 + \sum_{j=1}^q\Phi_j z_j$ is positive definite for all possible $z$'s. We then generate the discrete variables $\V z_i$'s based on $2^q$ probabilities given by $P_z$ in \eqref{moments}.    Since we use the exact probability rather than MCMC methods to generate the binary variable, the  memory requirements for the distribution $P_z$ makes it difficult to generate a large number of binary variables  in simulations.   However, this is not a problem for real data where the variables are already observed, and in fact our data analysis later in the paper demonstrates the method works well with large $q$.  Finally, for each $\V z_i$, we generate the continuous part $\V y_i$ from a multivariate Gaussian distribution with mean $\xi_{\V z_i}$ and covariance  $\Sigma_{\V z_i}$ defined by \eqref{moments}.


\begin{figure}[H]
  \centering
  \begin{subfigure}{\linewidth}
    \centering
    \includegraphics[width=1.05\linewidth]{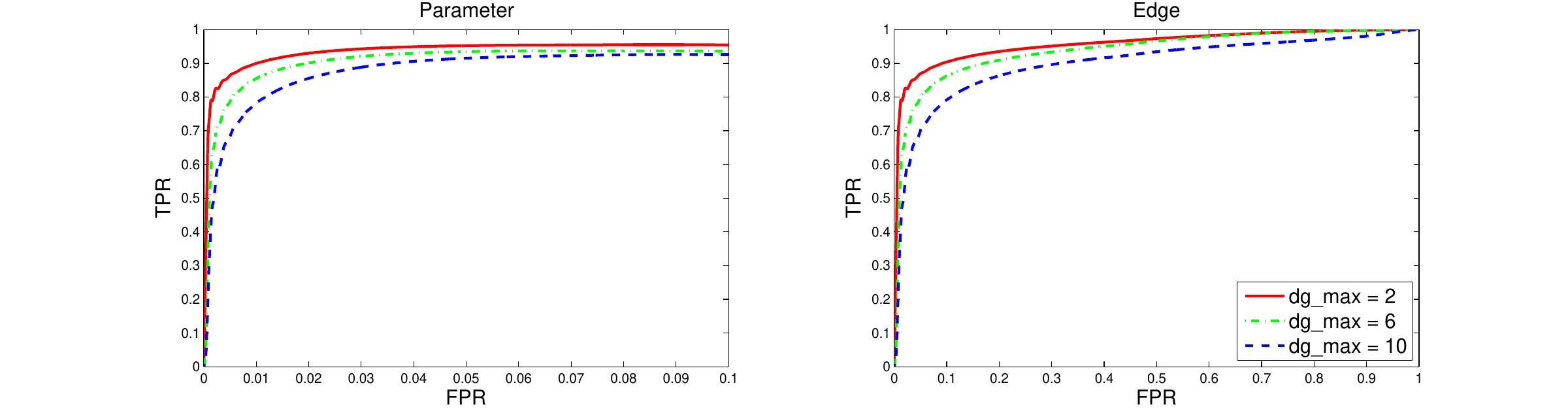}
    \caption{ROC curves for parameters and edges with varying maximum node degree. }
  \end{subfigure}
  \begin{subfigure}{\linewidth}
    \centering
    \includegraphics[width=1.05\linewidth]{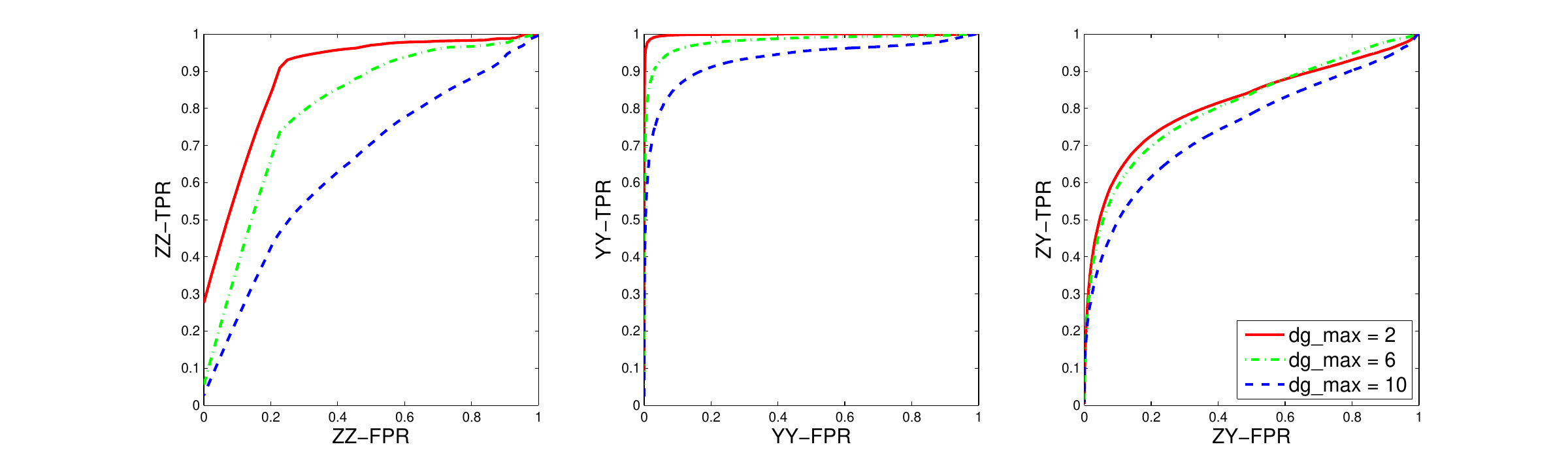}
    \caption{ROC curves for each edge category with varying maximum node degree.}
  \end{subfigure}  
  \caption{Model selection results. }
  \label{fig:marginal}
\end{figure}  
\vspace{-0.15in}

Figure~\ref{fig:marginal}(a) shows the impact of maximum node degree.  The more homogeneous the node degrees are, the easier the model selection task. This is because without prior information, the proposed method treats all nodes equally and uses the same tuning parameter for all regressions.   We also report ROC curves for each of the three edge types (ZZ, YY, ZY), shown in Figure~\ref{fig:marginal}(b). The pattern in each category is qualitatively consistent with the overall pattern in Figure~\ref{fig:marginal}(a), and it appears that accuracy on the edges between binary variables (ZZ) suffers the most from increased degree heterogeneity.

\subsection{Comparison with other graphical model methods}\label{subsec:simulation2}
Here we compare the proposed method with several other penalized regression approaches.   \citet{Fellinghauer11} proposed to fit separate $\ell_1$ regularized regressions by regressing each variable on the others, without including any interaction terms.  This is a special case of our model.  
 \citet{Lee12} fit the same model as \citet{Fellinghauer11} (no interaction terms) by maximizing a joint pseudo-likelihood instead of fitting separate regressions, and also applies calibration to adjust the penalty weights.   We also include a comparison to our model (with interaction terms) penalized by the regular lasso penalty instead of the weighted lasso penalty, effectively replacing the weight of $2$ in front of the intercation terms with $1$ while keeping everything else the same.    We implemented our method and the method of \citet{Fellinghauer11} using the glmnet package in matlab. The method of \citet{Lee12} is based on the matlab code provided by the authors.   The computational cost of our method is about the same as that of \citet{Fellinghauer11} and hundreds of times lower  than that of \citet{Lee12}.

We consider two simulation settings.  In the first setting, we set all $\Phi_j^{\gamma\mu}$, $j=1,\ldots, q; \gamma, \mu=1, \ldots, p$ parameters to zero.  Thus the true model is exactly what \citet{Fellinghauer11} and \citet{Lee12} assume (no interaction terms in regressions \eqref{logistic_regression} and \eqref{linear_regression}), and each edge is represented by a unique parameter: the edge corresponding to $(Z_j, Y_\gamma)$ is determined by $\eta_j^\gamma$, the edge for $(Z_j, Z_k)$ is determined by $\lambda_{jk}$, and the edge for $(Y_{\gamma}, Y_\mu)$ is determined by $\Phi_0^{\gamma\mu}$.   We follow the set-up of the simulation in Section \ref{secsec:performance}, setting the maximum node degree to $6$, the total number of edges to 125, the number of variables to $p=90$ continuous and $q=10$ categorical, the sample size to $n=100$, and $a = c = 1$ and $b = 2$. 

\begin{figure}[H]
\centering
\includegraphics[width=1\textwidth]{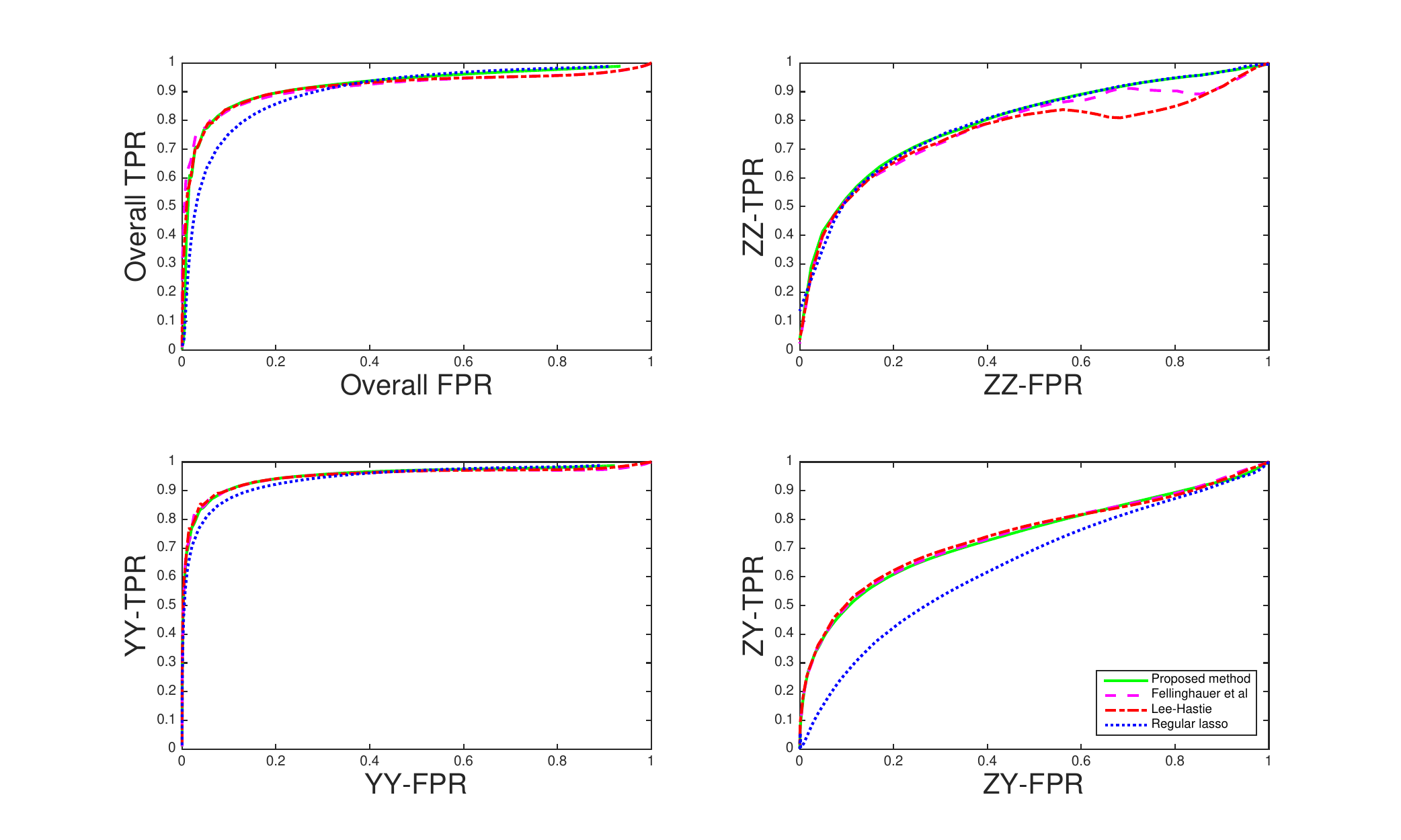}
\caption{Edge-based ROC curves for three graphical model methods when there are only main effects in the true model.}
\label{fig:comparison_model1}
\end{figure}

Figure \ref{fig:comparison_model1} shows that both \citet{Fellinghauer11} and \citet{Lee12} perform well, as is to be expected, since both \citet{Fellinghauer11} and \citet{Lee12} assume the true model.   Our method performs as well as \citet{Fellinghauer11} and \citet{Lee12}, meaning that it is able to learn that the interaction terms are irrelevant and recover the true model with main effects only.  The regular lasso penalty, on the other hand,  is inferior in this case, doing similarly on estimating $ZZ$ edges but worse on estimating $YY$ and $ZY$ edges. This likely happens because it penalized the interaction terms less than our method with the weighted lasso, and thus is not able to eliminate them as effectively.

In the second simulation setting, we allow for non-zero $\Phi_j^{\gamma\mu}$ parameters, keeping the dimensions $p=90$ and $q=10$ and the sample size $n=100$ fixed.  Note that, this corresponds to a graph that is sparse overall but has dense small locally dense subgraphs ($\Phi_j^{\gamma\mu} \neq 0$ indicates a $YZY$ clique).
 We first randomly generate 40 edges in the same way as in the first setting. Then we set $\{z_1, \cdots, z_4\}$, $\{z_8, \cdots, z_{10}, y_1, \cdots, y_6\}$ and $\{y_{11}, \cdots, y_{20}\}$ to be the three complete subgraphs, and there are no other edges in the graph.  The resulting graph has 127 edges, which is similar as before.  We also set the corresponding main and interaction effects in \eqref{logistic_regression} and \eqref{linear_regression} to be non-zero.    Here we set $a=0.1$, $b=0.2$, $c=0.6$ to obtain a reasonable signal-to-noise ratio and make sure the problem is neither impossible nor trivial.

\begin{figure}[H]
\centering
\includegraphics[width=1\textwidth]{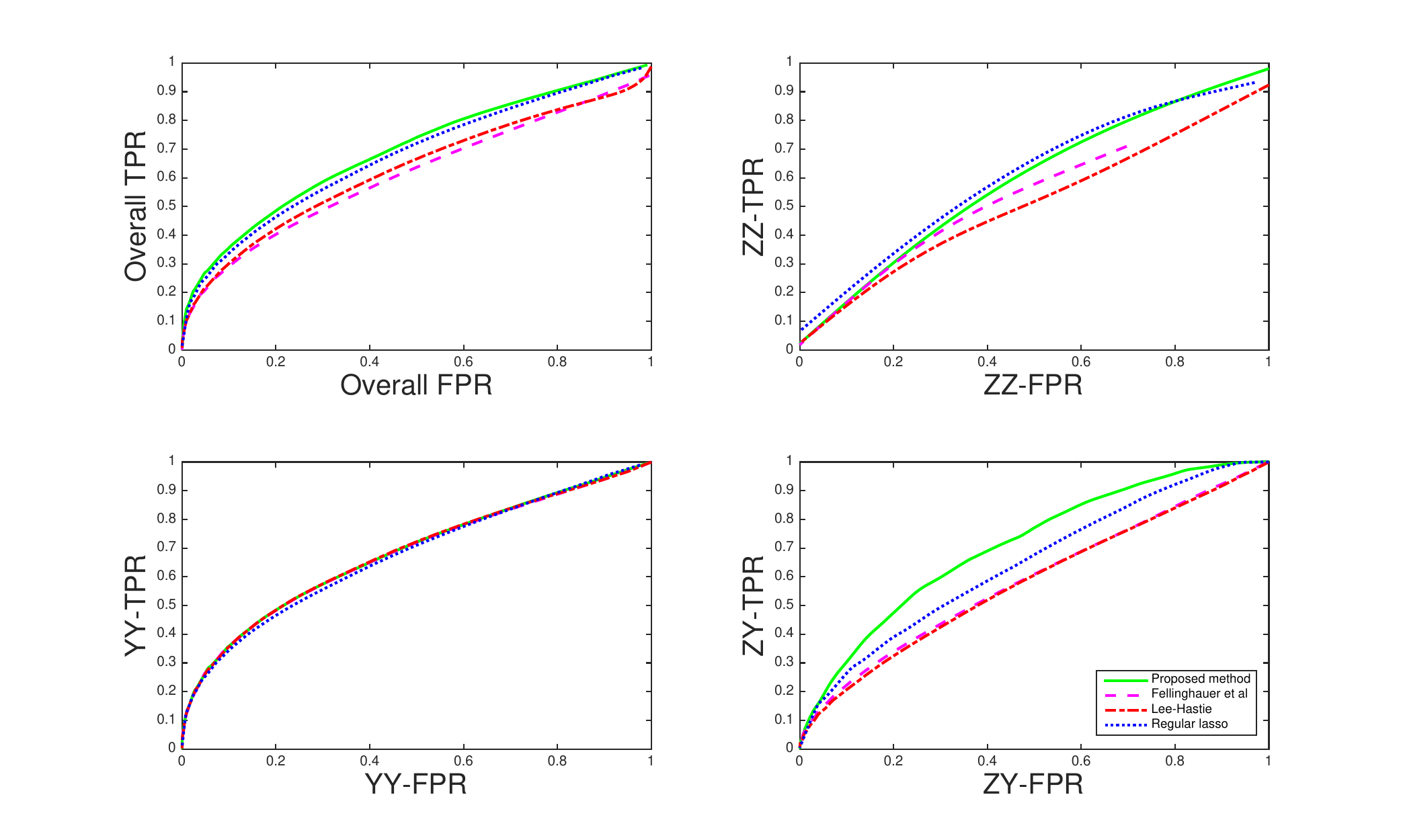}
\caption{Edge-based ROC curves for three graphical model methods when there are both main effects and interactions in the true model.}
\label{fig:comparison_model2}
\end{figure}

Figure~\ref{fig:comparison_model2} shows the results.  Since in this case \citet{Fellinghauer11} and \citet{Lee12} assume a wrong model, the comparison of the ROC curves for parameter identification is automically biased in our favor; instead, we show the ROC curves for edge identification only.  As expected, when interaction terms are present in the true model, our method outperforms both \citet{Fellinghauer11} and \citet{Lee12} on the overall ROC curve.  Decomposing the overall ROC curves into the three subtypes, we sees that there is no difference between the three methods on on the YY edges.  This may be because these edges involve only the continuous variables, and the Gaussian graphical model is often easier to estimate than the Ising model.   For edges involving discrete variables, which are more difficult to identify,  our method performs much better than \citet{Fellinghauer11} and \citet{Lee12} on ZY edges and somewhat better on the ZZ edges.  
The regular lasso, which, like our method, fits the true model here, gives results similar to our method on ZZ  and YY edges, and worse results on the ZY edges, resulting in a somewhat worse overall ROC curve.  

Overall, we observe that if the underlying model does not contain any interaction terms, both \citet{Fellinghauer11} and \citet{Lee12} perform well by not including them, and our model combined with the weighted lasso penalty does equally well by estimating these interactions to be 0.  When the true model does contain interaction terms, our method performs much better than \citet{Fellinghauer11} and \citet{Lee12} in terms of edge identification.   If we use our model with the regular lasso penalty instead of the weighted penalty, it performs a little worse when interaction terms are present but can be much worse when there are no interactions.

\begin{center}
\section{Application to music annotation} \label{sec:realdata}
\end{center}
\setcounter{section}{4}
\setcounter{subsection}{0}

Music annotation uses techniques from several disciplines, including audio signal processing, information retrieval, multi-label classification, and others.  Music annotation data sets usually consist of two parts: ``labels'', typically assigned by human experts, contain the categorical semantic description of the piece of music (emotions, genre, vocal type, etc.); and ``features'', continuous variables extracted from the time series of the audio signal itself using well-developed signal processing methods.   Representing these mixed  variables by a graphical model would allow us to understand how these different types of variables are associated with each other. For example, one can ask which rhythm and timbre features from the audio signal are associated with particular music genres, or emotions perceived to be conveyed by the music.  We apply our method to the publicly available music annotation data set \textit{CAL500} \citep{Turnbull08} from the \href{http://mulan.sourceforge.net/index.html}{Mulan} database \citep{Mulan} in order to find the conditional dependence patterns among these mixed variables.

\textit{CAL500} dataset consists of 502 popular music tracks (including songs with English lyrics and instrumental music) composed within the last 55 years by 502 different artists. The collection covers a large range of acoustic variations and music genres, and the labeling of each song is obtained from at least three individuals. 
For each song, the label part includes a semantic vocabulary of 149 tags represented by a 149-dimensional binary vector indicating the presence of each annotation.  These labels are partitioned into the following six categories:  emotions (36 total), genres (31), instruments (24), song characteristics (27), usages (15), and vocal types (16).  The continuous features are based on the short time Fourier transform (STFT) and are calculated for each short time window by sliding a half-overlapping, 23ms time window over the song's digital audio file. Detailed description of the feature extraction procedure can be found in \cite{Tzanetakis02}.  For each analysis window of 23ms, the following continuous features are extracted to represent the audio file: \textit{zero crossings}, a measure of the noisiness of the signal; \textit{spectral centroid}, a measure of `brightness' of the music texture with higher value indicating brighter music with more high frequencies; \textit{spectral flux}, a measure of the amount of local spectral change;  and the first MFCC coefficient \citep{Logan00} representing the amplitude of the music, which comes from a two-step transformation designed to capture the spectral structure. Every consecutive 512 of the 23ms short frames are then grouped into 1s long texture windows, based on which the following summary statistics for the four features defined above were calculated and used as  the final continuous variables: overall mean, mean of the standard deviations of each texture window, standard deviation of the means of each texture window, and standard deviation of the standard deviations of each texture window.

In our analysis, we omitted labels which were assigned to less than 3\% of the songs. Also, we standardized the continuous variables. This resulted in a dataset with $n = 502$ observations, $q = 118$ discrete variables, and $p = 16$ continuous variables.  

\begin{figure}[h!] 
\vspace{-3cm}
\includegraphics[width = 1.3\textwidth]{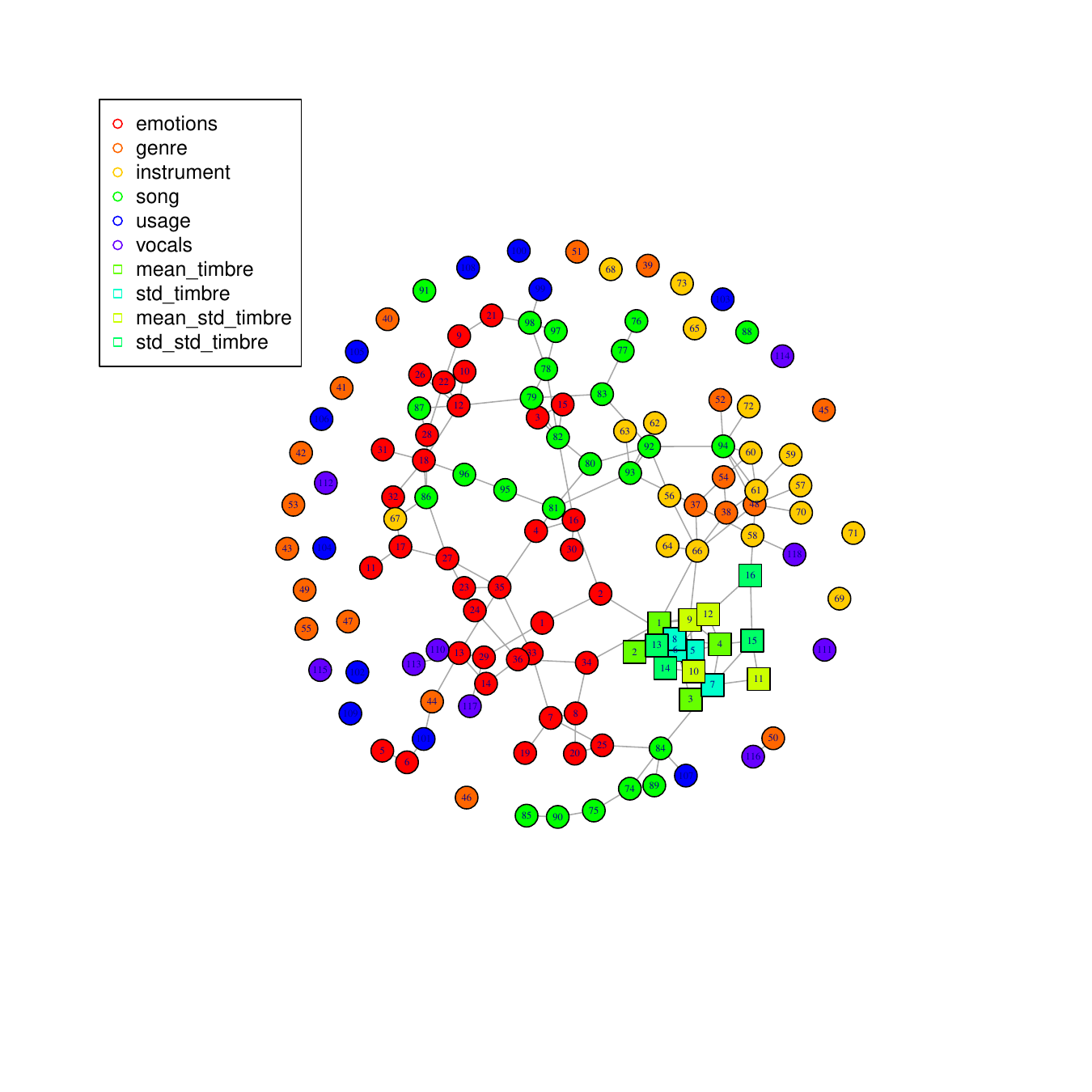}
\vspace{-4cm}
\caption{Estimated graphical model for CAL500 music data (edges with stability selection frequency of at least 99). }
\label{fig:music}
\end{figure}

We applied our method coupled with stability selection \citep{Meinshausen10} to identify the underlying graph for the purpose of exploratory data analysis, which is the primary usage of graphical models.  Stability selection was implemented by running the algorithm on 100 randomly drawn sub-samples of size $n/2$ for a grid of $(\rho, \kappa)$ values, and only keeping the edges that were selected at least 99\% of the time for a given value of $(\rho, \kappa)$.  
The results are shown in Figure \ref{fig:music}.  The continuous timbre features are represented by squares labeled 1-16 and the binary variables are represented by circles labeled 1-118. Each color represents a category of variables as shown in the legend. There are some interesting patterns within the group of binary labels, which allow us to infer connections between different emotions, genres, instruments, usages and so on. For example, the genre ``likable or popular songs" (circle 84) is associated with  ``catchy and memorable" (circle 74), ``would like to recommend" (circle 89) , the usage ``reading" (circle 107) and  the emotion ``pleasant and comfortable" (circle 25). Whether or not a song is ``very danceable" (circle 97, 98) is connected to ``fast tempo" (circle 78), the usage  ``at a party" (circle 99) and the emotion ``light and playful" (circle 21). We also find connections between the instrument ``piano" (circle 67),  and ``positive feelings" (circle 86),  ``not sad" (circle 32) and ``happy" (circle 17). The continuous variables that represent the audio signal features are quite densely connected within themselves, which is expected.  Edges connecting continuous and binary variables may also be interesting. For instance, the average noisiness of the music (square 1) is connected with emotions that are not tender or soft (circle 34) and emotions that are not angry or aggressive (circle 2) as well as with ``male lead vocals' (circle 66).


{\centering\section{Extension to general discrete data}\label{sec:extension}}

\setcounter{section}{5}
\setcounter{subsection}{0}

To extend our model to the general case where the discrete variables can take more than two values, we modify the previous model \eqref{model} into the following, \\
\begin{eqnarray}\displaystyle
\nonumber
\log f(z, y) &=& \sum_{d:d\subseteq\Delta, \lvert d\rvert \leq 2} \lambda_d(z) + \sum_{d:d\subseteq\Delta, \lvert d\rvert \leq 1} \eta_d(z)^Ty - \frac{1}{2}\sum_{d:d\subseteq\Delta, \lvert d\rvert \leq 1} y^T\Phi_d(z)y  \ , \\ 
\nonumber
&=& \left(\lambda_0 + \sum_{j=1}^q \lambda_j(z_j) + \sum_{j > k} \lambda_{jk}(z_j, z_k)\right) + \sum_{\gamma = 1}^p \left(\eta_0^{\gamma} + \sum_{j=1}^q \eta_j^{\gamma}(z_j) \right)y_{\gamma} \ , \\ 
\label{model_extension}
&& - \frac{1}{2}\sum_{\gamma, \mu = 1}^p\left(\Phi_0^{\gamma\mu} + \sum_{j=1}^q\Phi_j^{\gamma\mu}(z_j) \right)y_{\gamma}y_{\mu} \ ,
\end{eqnarray}
where each $z_j$ takes integer values 1 to $K_j$; $\lambda_j(\cdot), \ \eta_j^\gamma(\cdot), \ \Phi_j^{\gamma\mu}(\cdot)$ are all discrete functions which take on $K_j$ possible values and $\lambda_{jk}(\cdot, \cdot)$ is a discrete function with $K_j \times K_k$ values. For identifiability, we set $\lambda_j(1) = 0, \ \eta_j^\gamma(1) = 0,\  \Phi_j^{\gamma\mu}(1) =0$ and $\lambda_{jk}(1, \cdot) = \lambda_{jk}(\cdot, 1) = 0$.
The correspondence between the parameters and the edges is then given by 
\begin{eqnarray}
Z_j \perp Z_k \  \vert \ \ X \backslash \{Z_j, Z_k\} &\Longleftrightarrow & \V \theta_{jk} = \left( \lambda_{jk}(z_j, z_k)\right ) = 0  \nonumber \ , \\
Z_j \perp Y_{\gamma} \ \vert \ \ X \backslash\{Z_j, Y_\gamma\} &\Longleftrightarrow& \V \theta_{j\gamma} = \left(\eta_j^{\gamma}(z_j), \{\Phi_j^{\gamma\mu}(z_j): \mu \in \Gamma\backslash\{\gamma\}\} \right) = 0 \ ,  \nonumber \\ \label{edge_parameter_extension}
Y_{\gamma} \perp Y_{\mu} \ \vert \ \ X\backslash\{Y_\gamma, Y_\mu\} &\Longleftrightarrow& \V \theta_{\gamma\mu} = \left(\Phi_0^{\gamma\mu}, \{\Phi_j^{\gamma\mu}(z_j): j \in \Delta\}\right) =  0 \ .
\end{eqnarray}
The generalized model can be fitted with separate regressions based on the conditional likelihood of each variable. The parameters in \eqref{edge_parameter_extension}  still have a group structure, which calls for using the group lasso penalty as in \eqref{logistic_regression_l1l2} and \eqref{linear_regression_l1l2}. The structure of overlaps is more complex in this case, and we use the upper bound $\ell_1$ approximation as in \eqref{logistic_regression_l1} and \eqref{linear_regression_l1} to obtain the final estimates. Specifically, we minimize the following criteria separately:  \\

\noindent\textit{Logistic regression with $\ell_1$ penalty}: for $j = 1, \ldots , q$
\begin{equation*}\displaystyle
\min \ell_j + \rho \left( \kappa\sum_{k\neq j}\sum_{(z_j, z_k)}\vert\lambda_{jk}(z_j, z_k)\vert + \sum_{\gamma=1}^p\sum_{z_j = 1}^{K_j}\vert \eta_j^\gamma(z_j)\vert + 2\sum_{\gamma < \mu}\sum_{z_j = 1}^{K_j}\vert\Phi_j^{\gamma\mu}(z_j)\vert \right).  
\end{equation*}
\textit{Linear regression with $\ell_1$ penalty}: for $\gamma = 1, \ldots, p$
\begin{equation*} \displaystyle
\min \ \ell_\gamma +\rho \left( \sum_{j=1}^q\sum_{z_j = 1}^{K_j}\vert \tilde{\eta}_j^\gamma(z_j)\vert + \sum_{\mu \neq \gamma}\vert \tilde{\Phi}_0^{\gamma\mu} \vert + 2\sum_{j=1}^q\sum_{\mu \neq \gamma} \sum_{z_j=1}^{K_j}\vert \tilde{\Phi}_{j}^{\gamma\mu}(z_j)\vert\right).
\end{equation*}

 \citet{Yuan06} proposed further adjusting the weights in the group lasso penalty  for categorical variables to reflect its number of levels, which can carry over to our proposed weighted lasso 
approximation.

{\centering
\section{Discussion}\label{sec:summary}}
\setcounter{section}{6}
\setcounter{subsection}{0}

In this paper, we have proposed a new graphical model for mixed (continuous and discrete) data, which is particularly suitable for high-dimensional settings.     As discussed in the introduction, while the general conditional Gaussian model is well known and goes back to \citet{Lauritzen89}, it is not appropriate for high-dimensional data, and there is little previous work on mixed graphical models that can scale to modern applications.    Two recent new developments on this topic, \citet{Fellinghauer11} and \citet{Lee12},  were derived in parallel with and independently of this  manuscript.  Both \citet{Fellinghauer11} and \citet{Lee12} assume a more restricted version of the conditional Gaussian density by assuming constant conditional covariance for all the continuous variables and is thus a special case of our model \eqref{model}, where all the $\Phi_j$ are 0.    This can be too restrictive for some applications, since our model is the most parsimonious conditional Gaussian density that allows for varying conditional covariances, and we showed that when interaction terms are present in the model, our method does indeed perform much better.  
\citet{Fellinghauer11} considered fitting $\ell_1$-regularized regressions of each variable on the rest, 
while \citet{Lee12} considered the maximum pseudo-likelihood approach.
We chose to fit separate regressions, rather than maximize the joint pseudo-likelihood.  One reason is that the number of parameters in our model is $\mathcal{O}\left(\max(q^2, p^2q)\right)$, making maximizing the joint pseudo-likelihood computationally more expensive than in the simpler setting of  \citet{Lee12} with $\mathcal{O}\left(q^2+p^2\right)$ parameters; even in the simpler setting, maximizing the joint pseudo-likelihood is hundreds of times slower.   
Another reason is that we did not observe much difference between the separate regression approach and the joint pseudo-likelihood approach in the simpler setting of \citet{Fellinghauer11} and \citet{Lee12}.

As we already know from the literature on the Gaussian graphical model and the Ising model, estimating conditional independence relationships between binary variables is in general more challenging.  We observe it in this context as well, with interactions between two continuous variables being estimated better than interactions between two binary variables, or between a binary and a continuous variable.   Establishing theoretical performance guarantees for the overlapping group penalty is outside the scope of the present paper but presents an interesting challenge for the future, as the mixed variable setting is substantially more complicated than either the Gaussian or the pure binary setting.   However, with the weighted lasso penalty approximation the separate regressions we fit reduce to standard settings for either the lasso linear or the logistic regression, where model selection results under appropriate conditions have already been established.   We did not state these results in the paper since joint conditions for the continuous and the binary variables are awkward and require a lot of notation and space to write out,  the results themselves are standard, and this paper is not focused on theory;  nonetheless, this connection with the sparse regression literature guarantees reasonable behavior of our method provided the conditions are satisfied.  

\section*{Acknowledgements}

We thank the Associate Editor and two referees for their careful reading of the manuscript and many helpful suggestions.   This work  was performed when the first author was a PhD student at the University of Michigan.    E. Levina's  research was partially supported by NSF grants DMS-1106772, DMS-1159005, DMS-1521551;  J. Zhu's research was partially supported by NSF grant DMS-1407698 and NIH grant R01GM096194.  

\renewcommand{\refname}{\begin{center}REFERENCES\end{center}}
\bibliographystyle{asa}
\bibliography{allref}
 
\end{document}